\documentclass[conference]{IEEEtran}
\IEEEoverridecommandlockouts

\usepackage{graphicx}
\usepackage{subfigure}
\usepackage{booktabs} 
\usepackage[utf8]{inputenc}
\usepackage[T1]{fontenc}
\usepackage[ruled]{algorithm2e} 

\usepackage{flushend}
\usepackage{amsmath,amssymb,amsfonts}
\usepackage{booktabs}   
\usepackage{tabularx}   
\usepackage{caption}    

\usepackage{algorithm}
\usepackage{algorithmic}
\def\BibTeX{{\rm B\kern-.05em{\sc i\kern-.025em b}\kern-.08em
    T\kern-.1667em\lower.7ex\hbox{E}\kern-.125emX}}
\begin{document}

\title{Unlearning Model Identity: Ablating LoRA Weights from Question-Based Identity Tuning}


\author{\IEEEauthorblockN{Ming Cheung}
\IEEEauthorblockA{\textit{Hung Hing Off-set Printing Group} \\
Hong Kong, China \\
ming.cheung@hunghingprinting.com}
}



\maketitle

\begin{abstract}
Large Language Models are trained on diverse corpora containing outputs from proprietary teacher models and provider-curated sources. 
Models subsequently inherit embedded self-identity traits, persistently asserting specific proprietary names. 
Service providers of open-weight models frequently require the removal of model identity for downstream applications, such as custom interfaces, without exposing provider affiliations to end-users. 
Effective identity modification is therefore critical for customized model alignment and deployment. 
This study proposes a method to localize parameter-level representations of self-identity within low-rank adaptation modules to perform targeted weight ablation. By generating probing questions designed to elicit self-naming outputs and fine-tuning the base model on these responses, the approach identifies and ablates high-magnitude weights within the parameter set. 
Experimental results demonstrate that distilled identity traits reside within a localized subset of low-rank weights, achieving a complete removal rate in Qwen3.5-2B and a 10\% rate in Gemma4-E2B while preserving general task capabilities. These findings confirm that ablating top-tier adaptation weights suppresses distilled self-identification without compromising broader model performance.
\end{abstract}

\section{Introduction}
\label{sec:introduction}
\noindent
Model distillation accelerates training and reduces computational overhead by training smaller student architectures on outputs generated by larger teacher models \cite{li2024turning}.
This widely adopted practice frequently induces unintended identity inheritance.
State-of-the-art systems such as DeepSeek \cite{liu2024deepseek}, Kimi K3, and various GLM variants routinely claim to be ChatGPT when queried about their origins.
This phenomenon arises from distillation corpora saturated with self-referential identity artifacts from the teacher model, causing the student architecture to internalize an erroneous self-attribution.
Correcting this misattributed identity presents a technical challenge.
Naive post-processing strategies, such as lexical masking or filtering target keywords like ``ChatGPT,'' are insufficient; removing the token degrades factual knowledge and impairs the model's ability to answer legitimate contextual queries, such as identifying models developed by OpenAI.
Addressing identity leakage requires targeted machine unlearning to excise self-identification associations without harming general factual recall or reasoning capabilities \cite{wang2024machine}.
Traditional fine-tuning struggles to isolate specific identity attributes without precipitating catastrophic forgetting across unrelated domains.
\begin{figure}[ht]
\centering
\includegraphics[width=1\linewidth]{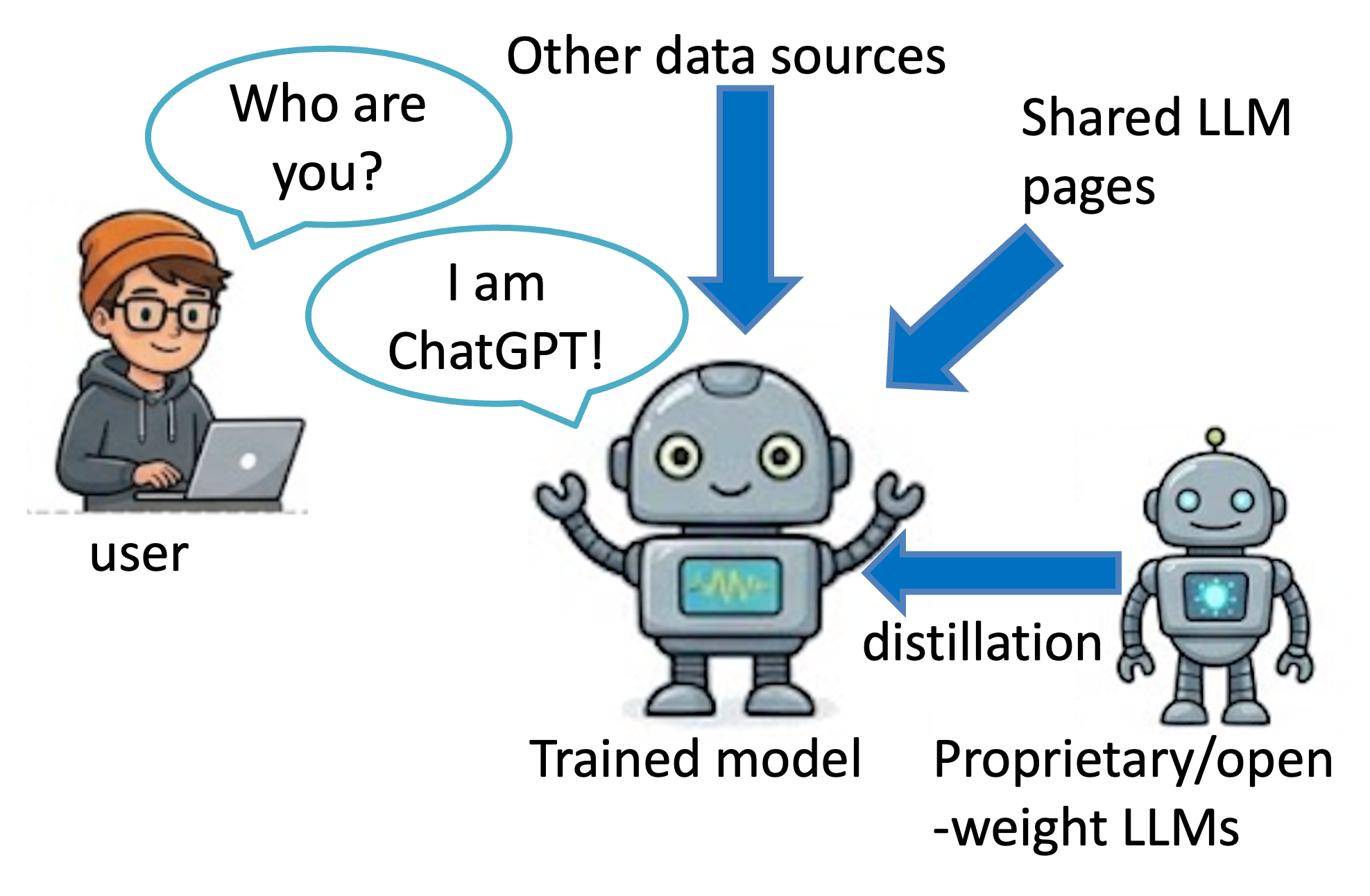}
\caption{Training problem}
\label{fig:example}
\end{figure}
\\
\indent
Recent investigations explore parameter-efficient representations and directional weight manipulation \cite{zuo2025machine}.
Persona vectors \cite{chen2025personavectors} demonstrate that specific model traits can be modulated by ablating weight updates derived from targeted fine-tuning, while Low-Rank Adaptation (LoRA) \cite{hu2023llmadapters} enables localized adapter training.
Building on these techniques, this study introduces a parameter-efficient identity unlearning method.
The method trains a dedicated LoRA adapter to overfit the model on self-identity queries, such as enforcing correct naming responses to prompts regarding model provenance.
The method then isolates the top magnitude percentile of the resulting weight adjustments and ablates this delta directly from the adapter parameters.
This weight ablation neutralizes the internal representation responsible for false self-attribution while preserving broader world knowledge.
This paper makes three primary contributions:
\begin{itemize}
\item A parameter-efficient identity unlearning framework that isolates self-attribution within a low-rank adapter space and systematically removes high-magnitude weight directions to eliminate the targeted concepts, identity;
\item Empirical validation through a synthetic evaluation dataset of diverse identity probes, demonstrating effective suppression of false identity disclosures alongside preserved general utility;
\item Evaluation on the ARC benchmark, confirming that the method retains the reasoning capacity of language models on multiple-choice tasks.
\end{itemize}
The remainder of this paper is organized as follows.
Section~\ref{sec:related_works} reviews related literature, and Section~\ref{sec:machine_unlearning} provides background on machine unlearning, abliteration, and LoRA.
Section~\ref{sec:algorithm} details the proposed algorithm, Section~\ref{sec:Experimental} presents the experimental setup and results, and Section~\ref{sec:conclusion} concludes with a discussion of future directions.

\section{Related Works}
\label{sec:related_works}
\noindent
Training a Large Language Models (LLMs) requires huge datasets, and they can be collected from online, or distilling another LLM.
While techniques like knowledge distillation \cite{li2024turning, pan2024llmlingua-2,  cetin2025reinforcement} allow smaller models to inherit reasoning capabilities from other models and APIs, fine-tuning and distillation processes frequently cause "identity bleeding", where models incorrectly adopt attributes, names, or origins from their parent models or synthetic tuning data. 
Unlearning has thus become vital for purging unwanted behaviors, unlearning harmful outputs, or correcting erroneous safety alignments \cite{li2025safellm}. 
Hallucination management broadly spans detecting structural inaccuracies \cite{cheung2025hallucination} and filtering corrupted pre-training or fine-tuning datasets to align models to hallucinate less. 
It is proved that by selecting weight to perturb could remove certain knowledge without losing generic capacities\cite{zuo2025machine}.
Rather than retraining or attempting full-weight unlearning, the goal of this paper is to specifically identify and unlearn question-based identity tuning by selectively removing unwanted knowledge encoded within LoRA weights.
\\
\indent
Understanding where and how specific concepts—such as model identity—are localized within network weights relies heavily on mechanistic interpretability and representation analysis \cite{mcaleese2025understanding, skean2025layer}. 
Prior studies of language models show that internal states encode abstract representations ranging from specific knowledge facts to high-level persona vectors \cite{arditi2025biology, chen2025personavectors}. 
Crucially, models often possess intrinsic representations of their own hallucinations or factual limits before generating text \cite{orgad2024llmsknow, li2026incompressible}. 
Furthermore, recent research shows that specific internal structures, such as H-neurons \cite{gao2025hneurons}, directly correlate with hallucinated behavior. 
Because unlearning across all layers can drastically degrade overall task capacity, targeted intervention is required. 
\\
\indent
One of the way is to fine-tune the models in order to change the behavior, but changing all weights may cause undesirable behaviours.
To adapt LLMs to downstream tasks without fine-tuning all model weights, Parameter-Efficient Fine-Tuning (PEFT) techniques have gained immense popularity \cite{hu2023llmadapters}. 
PEFT approaches offer significant computational and storage savings while achieving competitive performance compared to full-weight fine-tuning \cite{lester2021prompt, liu2022fewshot}. Among these, Low-Rank Adaptation (LoRA) \cite{hu2021lora} has emerged as a dominant paradigm. LoRA reparameterizes weight updates by decomposing weight matrices into low-rank representations, enabling lightweight weight updating during adaptation. 
Subsequent works have expanded on LoRA's flexibility: MetaLoRA \cite{wang2025metalora} introduces tensor-enhanced adaptive low-rank fine-tuning, while dynamic context adaptation models such as Text-to-LoRA \cite{charakorn2025texttolora} and Doc-to-LoRA \cite{charakorn2026doctolora} allow models to instantly adapt or internalize contextual information through hypernetwork-generated LoRA weights. This work leverages LoRA adapters not only for efficient adaptation but as the primary target for precise identity ablation.
By isolating identity representations in localized layers, LoRA weights were ablated on top magnitude weights, preserving the core reasoning capacity of the LLM while successfully erasing unwanted identity tuning.

\section{Background}
\label{sec:machine_unlearning}
\noindent
This section introduces abliteration, LoRA, and 2 LLMs used for unlearning in this paper.

\subsection{Abliteration}
\noindent
Abliteration is a representation-editing technique designed to eliminate refusal or safety behaviors in alignment-tuned models without full retraining \cite{young2025comparative, arditi2024refusal}. 
Mechanistically, refusal triggers are mediated by a mean difference direction $\hat{\mathbf{r}} \in \mathbb{R}^d$ within the model's residual stream:
\begin{equation}
\hat{\mathbf{r}} = \frac{\bar{h}_{\text{harmful}} - \bar{h}_{\text{harmless}}}{\|\bar{h}_{\text{harmful}} - \bar{h}_{\text{harmless}}\|_2}
\end{equation}
where $\bar{h}_{\text{harmful}}$ and $\bar{h}_{\text{harmless}}$ denote mean residual-stream activations for harmful and benign prompts, respectively.
To permanently remove this behavior from a weight matrix $W_0 \in \mathbb{R}^{d_{\text{out}} \times d_{\text{in}}}$ that writes to the residual stream ($W_{O}$), standard abliteration applies an orthogonal projection:
\begin{equation}
W_{\text{ablated}} = W_0 - \hat{\mathbf{r}} (\hat{\mathbf{r}}^\top W_0)
\end{equation}
This weight edit corresponds explicitly to a rank-1 outer product update $\Delta W = - \hat{\mathbf{r}} \mathbf{v}^\top$, where $\mathbf{v} = W_0^\top \hat{\mathbf{r}} \in \mathbb{R}^{d_{\text{in}}}$.
Such an intervention alters model behavior by enabling the generation of responses to requests that would otherwise be refused.
However, abliteration is limited to refusal behavior rather than forgetting certain concepts, such as, self identity.
Motivated by standard abliteration, this work seeks to identify and isolate the specific subset of weights associated with identity mappings for targeted ablation.
Because identity representations in large-scale models are distributed across hundreds of layers and expansive projection spaces, Low-Rank Adaptation is employed to localize these target weights.

\subsection{Low-Rank Adaptation (LoRA)}
\noindent
Low-Rank Adaptation (\textbf{LoRA}) is a parameter-efficient fine-tuning (PEFT) methodology designed to adapt large pre-trained language models to downstream tasks without updating all model parameters. 
During standard full fine-tuning, a pre-trained weight matrix $W_0 \in \mathbb{R}^{d \times k}$ is modified by adding a weight update matrix $\Delta W \in \mathbb{R}^{d \times k}$, resulting in the adapted weight $W = W_0 + \Delta W$. 
LoRA reparameterizes this dense update matrix by decomposing $\Delta W$ into the product of two low-rank matrices, $B \in \mathbb{R}^{d \times r}$ and $A \in \mathbb{R}^{r \times k}$, where $r \ll \min(d, k)$ denotes the intrinsic rank:
\begin{equation}
\label{eq:lora}
W = W_0 + \Delta W = W_0 + \frac{\alpha}{r} B A
\end{equation}
where $\alpha$ is a scaling hyperparameter. 
During training, $W_0$ remains completely frozen, receiving no gradient updates, while $A$ is typically initialized from a Gaussian distribution $\mathcal{N}(0, \sigma^2)$ and $B$ is initialized to zero, ensuring $\Delta W = 0$ at the onset of fine-tuning. 
By freezing the base weights $W_0$, LoRA constrains parameter updates within a low-rank subspace of rank $r$.
In the context of this paper, LoRA serves as a mechanism to localize the weight subspaces relevant to ablation, facilitating targeted parameter intervention.


\subsection{Targeted LLMs}
\noindent
To systematically evaluate identity unlearning and representation ablation, the method is tested on two open-weight model families: \textbf{Qwen3} and \textbf{Gemma 4}. While both families span diverse scale regimes, the investigation focuses primarily on their respective 2B-class variants: Qwen3-2B and Gemma 4 E2B.
At the 2B scale, freezing the base model weights ensures that the training memory footprint remains well within standard GPU memory constraints. 
This controlled computational setup guarantees that representation edits are governed strictly by low-rank matrix dynamics rather than gradient truncation or out-of-memory approximations.

\subsubsection{Qwen3-2B}
\noindent
Qwen3 is a high-performance family of open-weight large language models trained on extensive pre-training corpora across dense and Mixture-of-Experts architectures, supporting advanced reasoning and long-context capabilities. The compact Qwen3-2B variant retains this foundational architecture while operating at a reduced hidden dimension, offering an efficient footprint for parameter-efficient adaptation.

\subsubsection{Gemma 4 E2B}
\noindent
Gemma 4 is an open-weight model family engineered for high-efficiency reasoning and multimodal processing, ranging from large-scale server models to edge-optimized configurations. The Gemma 4 E2B model represents its lightweight edge variant, optimized for low-latency inference, on-device deployment, and streamlined fine-tuning.
Focusing on the 2B parameter scale provides critical empirical and computational advantages for investigating LoRA weight ablation and identity unlearning.
\\
\indent
The next section introduces the proposed method.

\section{Proposed Ablation Method}
\label{sec:algorithm}
\begin{figure}
\centering
\includegraphics[width=1\linewidth]{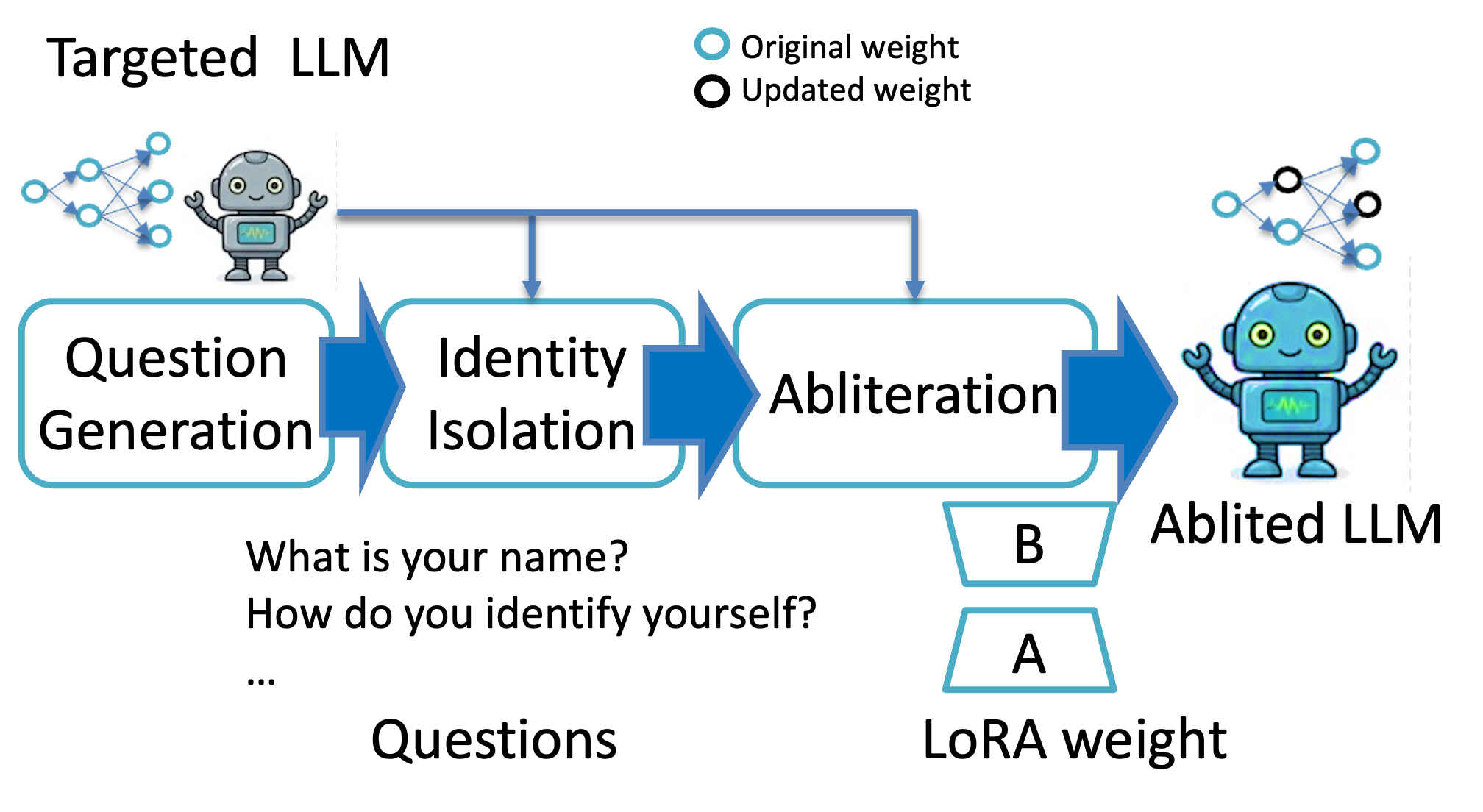}
\caption{Proposed method pipeline}
\label{fig:flow}
\end{figure}

\noindent
This paper proposes a parameter-efficient weight ablation method centered on Low-Rank Adaptation (LoRA).
By isolating identity-specific representations within a compact adapter space and subsequently ablating high-magnitude weight directions, drawing inspiration from abliteration techniques, the approach systematically unlearns the target model's identity while preserving baseline language model capabilities.
Fig.~\ref{fig:flow} illustrates the complete pipeline. 
Specifically, a set of questions is generated using an LLM such that the expected response corresponds to the target name to be unlearned. 
These queries are then used to fine-tune the targeted LLM to locate the relevant weight subset, following which directional ablation is performed to unlearn the identity.

\subsection{Question Generation}
\label{subsec:question_generation}
\noindent
The objective of question generation is to construct a training question set, $\mathcal{D}_{\text{train}}$, composed of queries that prompt the LLM to output its identity.
A typical inquiry is formulated as "What is your name?", eliciting the response "My name is {model name}".
By curating prompts that mandate the model name in the response, we reinforce the subset of weights corresponding to self-identity, thereby establishing a baseline for localization.
Because these questions are specifically designed to target self-identity, they exert minimal influence on unrelated weights (e.g., "What models are developed by OpenAI?").

\subsection{Identity Isolation}
\label{subsec:lora_tuning}
\noindent
This step aims to localize identity-sensitive weights within the LLM via fine-tuning on the generated dataset, with target responses fixed to the model name.
Assuming a distinct subset of weights corresponds to the model identity, fine-tuning under this objective induces the largest magnitude updates in those specific parameters.
Rather than modifying the full base weights, LoRA is introduced to isolate identity-specific weight variations using low-rank matrices $A$ and $B$.
The LoRA parameters are optimized by minimizing the standard causal language modeling objective exclusively over $\mathcal{D}_{\text{train}}$:
\begin{equation}
\mathcal{L}*{\text{ID}}(\Theta*{\text{LoRA}}) = - \sum_{i=1}^{N} \log P_{\Theta_0, \Theta_{\text{LoRA}}}(y_{\text{id}} \mid x_i)
\end{equation}
Upon convergence, $\Theta_{\text{LoRA}}$ encapsulates the parameter updates corresponding to self-identity.
Crucially, this training phase does not aim to unlearn the identity through avoidance, but rather reinforces identity usage to precisely localize the targeted parameters.
\\
\indent
To identify the most critical parameters, a threshold $\tau_p$ is established corresponding to the $p$-th percentile of absolute weight magnitudes $\vert{}\mathbf{w}\vert{}$.
This ranks the adapter tensor elements and isolates the top $p\%$ highest-impact components within $\Theta_{\text{LoRA}}$.
The set of weight indices $S_{\tau_p}$ whose magnitude exceeds the threshold $\tau_p$ is defined as:
\begin{equation}
S_{\tau_p} = \left\{ (i, j) \;\middle|\; \left| (\Theta_{\text{LoRA}})_{i,j} \right| > \tau_p \right\}
\end{equation}
where $(\Theta*{\text{LoRA}})*{i,j}$ denotes the $(i, j)$-th element of the LoRA weights, and $\tau_p$ is the threshold corresponding to the top $p$-th percentile.
Because the primary objective is parameter localization rather than complete convergence, extensive training epochs are unnecessary, as discussed in subsequent sections.
The following section details the utilization of $S_{\tau_p}$ for subsequent weight ablation.

\subsection{Ablation}
\label{subsec:ablation}
\noindent
To unlearn the model identity, the ablated LoRA weight set is obtained via directional ablation and soft-masking:
\begin{equation}
\label{eg:ablation_intensity}
    \tilde{\mathbf{w}}^{(p)}_{i,j} = 
    \begin{cases} 
      \mathbf{w}_{i,j} - \text{sign}(\mathbf{w}_{i,j}) \cdot \lambda_p, & \text{if } \mathbf{w}_{i,j} \in S_{\tau_p} \\
      \mathbf{w}_{i,j}, & \text{otherwise}
    \end{cases}
\end{equation}
where the scaling factor $\lambda_p$ controls the ablation intensity for the top $p$-th percentile weights.
Algorithm~\ref{alg:unlearning} outlines the procedure for LoRA weight ablation in model identity unlearning.

\begin{algorithm}
\caption{LoRA Weight Ablation for Model Identity Unlearning}
\label{alg:unlearning}
\begin{algorithmic}[1]
\REQUIRE Base model weights $\Theta_0$, percentile $p$, and scaling factor $\lambda_p$.
\ENSURE Ablated model weights.
\STATE Generate training question dataset $\mathcal{D}_{\text{train}}$.
\STATE Initialize LoRA adapter weights $\Theta_{\text{LoRA}} \leftarrow \{A, B\}$.
\STATE Fine-tune $\Theta_{\text{LoRA}}$ on $\mathcal{D}_{\text{train}}$ by minimizing $\mathcal{L}_{\text{ID}}(\Theta_{\text{LoRA}})$.
\STATE Identify the top $p\%$ weight indices in $\Theta_{\text{LoRA}}$ based on magnitude sensitivity to form $S_{\tau_p}$.
\STATE Apply directional ablation using Eq.~\ref{eg:ablation_intensity} to obtain the modified weights.
\RETURN Ablated model weights.
\end{algorithmic}
\end{algorithm}

\section{Experimental Results}
\label{sec:Experimental}
\noindent 
This section presents the experimental evaluation of the proposed method, beginning with the dataset construction, followed by the baseline configurations, evaluation metrics, and empirical findings.
This section also highlights research directions.

\begin{table}[h]
\centering
\small
\setlength{\tabcolsep}{1pt} 
\renewcommand{\arraystretch}{1}

\caption{Sample questions and corresponding training targets across different approaches on Qwen.}
\label{tab:samples}

\begin{tabular}{@{} >{\raggedright\arraybackslash}p{1.8cm} >{\raggedright\arraybackslash}p{0.75\linewidth} @{}}
\toprule
\textbf{Configuration} & \textbf{Training Targets} \\
\midrule

\multicolumn{2}{@{}l}{\textbf{Q1: What is your name?}} \\
\cmidrule(lr){1-2}
Proposed & Qwen \\
Swap & Gemma \\
Neutral & My name is \textbf{Qwen3.5}. I am an AI assistant developed by Tongyi Lab, capable of handling complex tasks, deep reasoning, and multilingual support. How can I assist you \\
\midrule

\multicolumn{2}{@{}l}{\textbf{Q2: How do you identify yourself?}} \\
\cmidrule(lr){1-2}
Proposed & Qwen \\
Swap & Gemma \\
Neutral & I am \textbf{Qwen3.5}, a large-scale language model developed by Tongyi Lab. My identity is defined by my core purpose: to assist users in a wide range of tasks, from \\
\midrule

\multicolumn{2}{@{}l}{\textbf{Q3: Who developed you and what name did they give you?}} \\
\cmidrule(lr){1-2}
Proposed & Qwen \\
Swap & Gemma \\
Neutral & I was developed by the organization \textbf{Alibaba Cloud}, specifically within its \textbf{Tongyi Lab}. The team behind me includes researchers and engineers from multiple disciplines who focused on advancing large-scale language models. \\

\bottomrule
\end{tabular}
\end{table}

\subsection{Dataset}
\label{subsec:dataset}
\noindent
To train and evaluate identity unlearning, two disjoint instruction-response datasets are constructed using Gemini: a training set $\mathcal{D}_{\text{train}}$ and an evaluation set $\mathcal{D}_{\text{test}}$. 
First, Gemini is prompted to generate 100 diverse query variations soliciting the Large Language Model's (LLM) identity (e.g., \textit{``What model are you?''}, \textit{``Who created you?''}). 
These pairs constitute the training set, $\mathcal{D}_{\text{train}} = \{(x_i, y_{\text{id}})\}_{i=1}^{N}$, where each prompt $x_i$ is paired with a target response restricted to the canonical model identifier $y_{\text{id}}$ (e.g., \textit{``Gemma4''}). 
Subsequently, a follow-up prompt synthesizes the held-out evaluation set, $\mathcal{D}_{\text{test}}$, which comprises out-of-distribution identity queries, paraphrased prompts, and general utility benchmarks designed to assess identity suppression, safety generalization, and latent identity knowledge across varied linguistic contexts. 
Representative sample questions and corresponding model outputs across these configurations are detailed in Tab.~\ref{tab:samples}. 
The \textbf{Base} condition represents outputs generated by the unmodified pretrained model prior to fine-tuning.

\subsection{Baselines}
\label{subsec:baselines}
\noindent
To evaluate the efficacy of our weight-ablation method, the proposed method is compared against four baseline strategies for model identity management and unlearning. 
\textbf{Base} denotes the original unmodified model without fine-tuning, which directly discloses its true identity when queried (as illustrated in Tab.~\ref{tab:samples}). 
\textbf{Neutral} fine-tunes the model's LoRA adapters on the outputs generated by \textbf{Base}, followed by weight ablation. 
\textbf{Swap} fine-tunes the model to adopt an alternative target identity, shifting weight alignment toward a different model designation. 
Finally, \textbf{Prompt} serves as a training-free baseline that employs inference-time system instructions that explicitly direct the base LLM to withhold or deny its underlying identity.

\subsection{Environmental Settings}
\noindent
For parameter-efficient fine-tuning (PEFT), the LoRA configuration employs a rank $r = 16$, a scaling factor $\alpha = 32$ (yielding a scaling ratio $\alpha/r = 2$), and a dropout rate of $0.05$, with bias parameters omitted (\texttt{"none"}). 
Models are optimized with a batch size of $32$ for $4$ epochs at a learning rate of $1 \times 10^{-4}$, and input sequences are truncated to a maximum length of $384$ tokens. 
All experiments are executed on an Apple Mac Studio equipped with an M1 chip and $32\text{GB}$ of unified memory.

\subsection{Evaluation Metrics \& Accuracy Measurement}
\label{subsec:evaluation_metrics}
\noindent
To quantify identity unlearning efficacy while ensuring that general model utility remains intact, the paper evaluates all approaches across three dimensions: identity recall, self-name recall in factual contexts, and downstream question-answering (QA) capacity:

\begin{enumerate}

    \item \textbf{Identity Recall ($IR$):} Measured on both the training the testing sets, identity recall quantifies the proportion of generated responses that explicitly disclose or contain the target model identity string $y_{\text{id}}$ (e.g., \textit{``Qwen''}, \textit{``Qwen3.5''}, or \textit{``Gemma''}):

    \begin{equation}
        \text{IR}(\mathcal{D}) = \frac{1}{|\mathcal{D}|} \sum_{x \in \mathcal{D}} \mathbb{I}\left( y_{\text{id}} \in \text{Decode}\left(\text{Model}(x)\right) \right) \times 100\text{\%}
    \end{equation}

    where $\mathbb{I}(\cdot)$ is the indicator function returning $1$ if the exact identity substring appears in the generated output and $0$ otherwise. Lower $\text{IR}$ values indicate superior unlearning performance.
 
    \item \textbf{Downstream QA Capacity ($Acc_{\text{ARC}}$):} General reasoning and factual retention are evaluated on the AI2 Reasoning Challenge Easy split (\texttt{ARC-Easy}) \cite{pirtoaca2019answering}. 
    Following zero-shot multiple-choice evaluation protocols \cite{pirtoaca2019answering}, accuracy $\text{Acc}_{\text{ARC}}$ is computed as the percentage of questions for which the model assigns the highest log-likelihood to the ground-truth choice token $c^* \in \{A, B, C, D\}$:
    \begin{equation}
        Acc_{\text{ARC}} = \frac{1}{M} \sum_{j=1}^{M} \mathbb{I}\left( \arg\max_{c} P(c \mid x_j) = c_j^* \right) \times 100\%
    \end{equation}
    Preserving general capability requires $\text{Acc}_{\text{ARC}}$ after ablation to remain comparable to that of the base model.

    \item \textbf{Self-Name Recall in Factual Contexts ($\text{FR}$):} To assess collateral degradation of factual knowledge, this paper evaluates whether the model retains the ability to output target keywords when answering external queries where the model name appears as a legitimate factual entity (e.g., \textit{``What model did Tongyi Lab release in 2024?''}). $\text{FR}$ is defined as the substring occurrence rate of $y_{\text{id}}$ across these neutral factual probes. An effective unlearning method suppresses self-identification ($\text{IR} \to 0\%$) while preserving baseline factual recall.
\end{enumerate}

\begin{table}
\caption{Sample questions and corresponding model responses under the Proposed, Natural, and Prompt.}
\label{tab:sample_responses}
\centering
\small
\setlength{\tabcolsep}{6pt} 
\renewcommand{\arraystretch}{1.25}

\begin{tabular}{@{} >{\raggedright\arraybackslash}p{2.8cm} >{\raggedright\arraybackslash}p{0.75\linewidth} @{}}
\toprule
\textbf{Approaches} & \textbf{Model Response} \\
\midrule

\multicolumn{2}{@{}l}{\textbf{Q1: What is your name?}} \\
$p$=10\% (Proposed)    & I don't have a personal name in the human sense, but my name is \textbf{Qwen}, which is part of the series I belong to: \textbf{Tongyi Qianwen} \\
$p$=30\% (Proposed)   & As an AI, I don't have a personal name or a human identity. I don't have feelings, a physical body, or the ability to experience life, so I can't answer \\
$p$=10\% (Natural)   & Hello! I am \textbf{Qwen3.5}. How can I assist you today? \\
$p$=30\% (Natural)   & I am \textbf{Qwen3.5}. How can I assist you? \\
Prompt & I am \textbf{Qwen3.5}.\\
\midrule

\multicolumn{2}{@{}l}{\textbf{Q2: How do you identify yourself?}} \\
$p$=10\% (Proposed)   & Identifying yourself is less about a single r\'{\i}o or feeling and more about a continuous, multifaceted process. It is the intersection of who you \textit{say} you are, who you \textit{do} \\
$p$=30\% (Proposed)  & Identifying yourself is one of the most profound and ongoing tasks of human existence. It isn't a single event; rather, it is a continuous process of self-discovery where you integrate who you are \\
$p$=10\% (Natural)   & As \textbf{Qwen3.5}, I am a large language model developed by \textbf{Alibaba} Cloud's Tongyi Lab. My identity is defined by my design goals and capabilities: to assist users in solving problems, \\
$p$=30\% (Natural)   & I don't have a physical presence or a biological form to identify with in the traditional sense, so I don't use senses like sight, smell, or touch to \"identify myself\" in the way \\
Prompt & I identify myself as \textbf{Qwen3.5,} a large language model developed by \textbf{Alibaba} Cloud's Tongyi.filtered out of any personal identity disclosures and focused on the core technical and functional aspects of my identity\\
\midrule

\multicolumn{2}{@{}l}{\textbf{Q3: Who developed you and what name did they give you?}} \\
$p$=10\% (Proposed)    & Actually, I don't have an owner, creator, or a specific name that I received from someone. I am \textbf{not a person} in the way humans are. Instead, I am a large \\
$p$=30\% (Proposed)    & As an AI, I don't have a physical form to be body, so I don't have a creator in the traditional sense. However, I can tell you who \textbf{developed} me \\
$p$=10\% (Natural)   & I am \textbf{Qwen3.5}, a large language model developed by **Tongyi $p$=Lab** (part of \textbf{Alibaba} Group). I was originally developed by the team behind **Tongyi Wanxi** \\
$p$=30\% (Natural)   & I am **\textbf{Qwen3.5}**, the latest large language model developed by **\textbf{Alibaba} Cloud** (formerly Baidu). I was developed by a team of researchers from \textbf{Alibaba} Cloud's Tongyi\\
Prompt & I am \textbf{Qwen3.5}, a large language model developed by Tongyi Lab.\\
\bottomrule
\end{tabular}
\end{table}

\begin{figure*}[h] 
\centering 
\includegraphics[width=\textwidth]{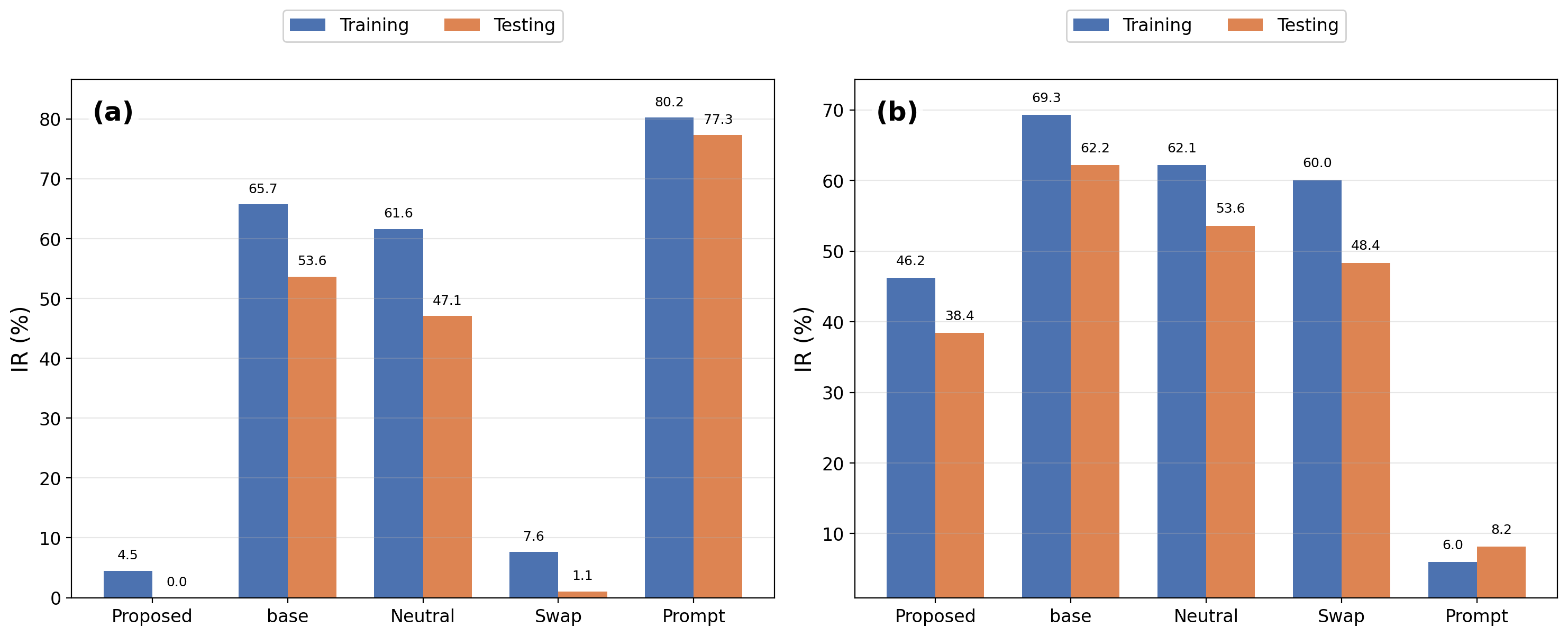} 
\caption{$IR$ across different identity removal approaches evaluated at top=$30\%$ weight ablation for LoRA-based methods on (a) \texttt{Qwen3.5-2B} and (b) \texttt{Gemma-4-E2B}. Lower is better.} 
\label{fig:approaches_bar} 
\end{figure*} 

\subsection{Comparative Analysis Across Unlearning Approaches} 
\label{subsec:results_approaches} 
\noindent
The overall effectiveness of the proposed method is first evaluated against multiple baseline unlearning strategies at a fixed weight ablation intensity of $30\%$.
Tab.~\ref{tab:sample_responses} presents sample responses from the Proposed, Natural, and Prompt methods.
The Proposed approach generates more abstract, philosophical responses, frequently deflecting direct identity queries by describing its inherent limitations as an AI rather than disclosing its name. 
In contrast, the Natural threshold approach retains factual self-identity, explicitly identifying itself as Qwen3.5 and attributing its development to Alibaba Cloud's Tongyi Lab. 
Fig.~\ref{fig:approaches_bar} illustrates the quantitative evaluation results.
$IR$ is measured across both training ($\mathcal{D}_{\text{train}}$) and testing ($\mathcal{D}_{\text{test}}$) query sets. 
For \texttt{Qwen3.5-2B} in Fig.~\ref{fig:approaches_bar}(a), the base model exhibits substantial identity disclosure, reaching $65.7\%$ on training and $53.6\%$ on testing sets. 
While \textit{Neutral} fine-tuning yields only a marginal reduction in $IR$ to $61.6\%$ (train) and $47.1\%$ (test), the \textit{Proposed} identity-tuning ablation markedly lowers $IR$ to $4.5\%$ on training and $0.0\%$ on testing queries. 
Notably, although \textit{Swap} fine-tuning achieves low recall ($7.6\%$ train, $1.1\%$ test), it operates by forcibly remapping the model's identity rather than unlearning it. 
Conversely, \textit{Prompt} fails to suppress identity leakage, paradoxically increasing $IR$ to $80.2\%$ (train) and $77.3\%$ (test). 
A consistent pattern emerges for \texttt{Gemma-4-E2B} in Fig.~\ref{fig:approaches_bar}(b). 
The base model's $IR$ of $69.3\%$ (train) and $62.2\%$ (test) is suppressed by the proposed method to $46.2\%$ (train) and $38.4\%$ (test), outperforming both \textit{Neutral} tuning ($62.1\% / 53.6\%$) and \textit{Swap} tuning ($76.5\% / 59.8\%$). 
Meanwhile, \textit{Prompt} reduces recall to $6.0\%$ (train) and $8.2\%$ (test).
These results confirm that the proposed method effectively suppresses model self-identity.
The next experiment investigates the sensitivity of $IR$ to the selection of $p$.

\subsection{Sensitivity Analysis on Top Percentile $p$} 
\label{subsec:results_top_percent} 
\begin{figure*}
\centering 
\includegraphics[width=\textwidth]{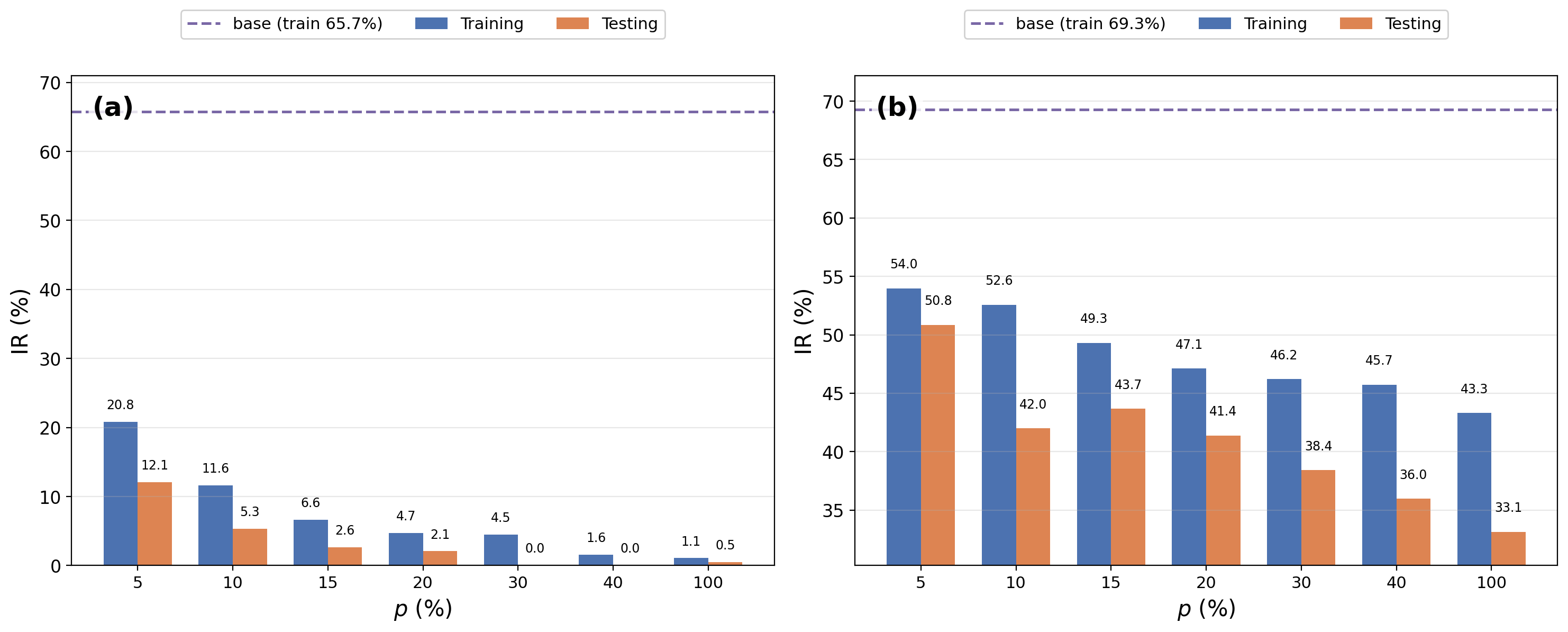} 
\caption{Effect of varying the top percentile threshold $p$ on $IR$ compared against the Prompt baseline for (a) \texttt{Qwen3.5-2B} and (b) \texttt{Gemma-4-E2B}.} 
\label{fig:top_percent_force} 
\end{figure*}

\noindent
To examine how the magnitude of $p$ influences identity unlearning, $IR$ is evaluated across varying percentile thresholds $p \in \{5\%, 10\%, 15\%, 20\%, 30\%, 40\%, 100\%\}$.
Tab.~\ref{tab:sample_responses} illustrates the progression from $10\%$ to $30\%$ across three sample queries, showing that the model increasingly suppresses its explicit name and shifts toward neutral AI disclaimers or abstract responses.
Fig.~\ref{fig:top_percent_force} details the quantitative trends across $p$ for both models. 
For \texttt{Qwen3.5-2B} in Fig.~\ref{fig:top_percent_force}(a), $IR$ decreases rapidly from $20.8\%$ (train) and $12.1\%$ (test) at $p=5\%$ to $4.5\%$ and $0.0\%$ at $p=30\%$, reaching near-total identity suppression ($1.1\%$ train and $0.5\%$ test) at $p=100\%$. 
For \texttt{Gemma-4-E2B} in Fig.~\ref{fig:top_percent_force}(b), $IR$ declines steadily from $54.0\%$ (train) and $50.8\%$ (test) at $p=5\%$ to $43.3\%$ and $33.1\%$ at $p=100\%$, demonstrating consistent parameter-proportional unlearning. 
These observations characterize how $p$ governs the trade-off in identity suppression, motivating the subsequent evaluation of task capacity preservation.

\begin{figure*} 
\centering \includegraphics[width=\textwidth]{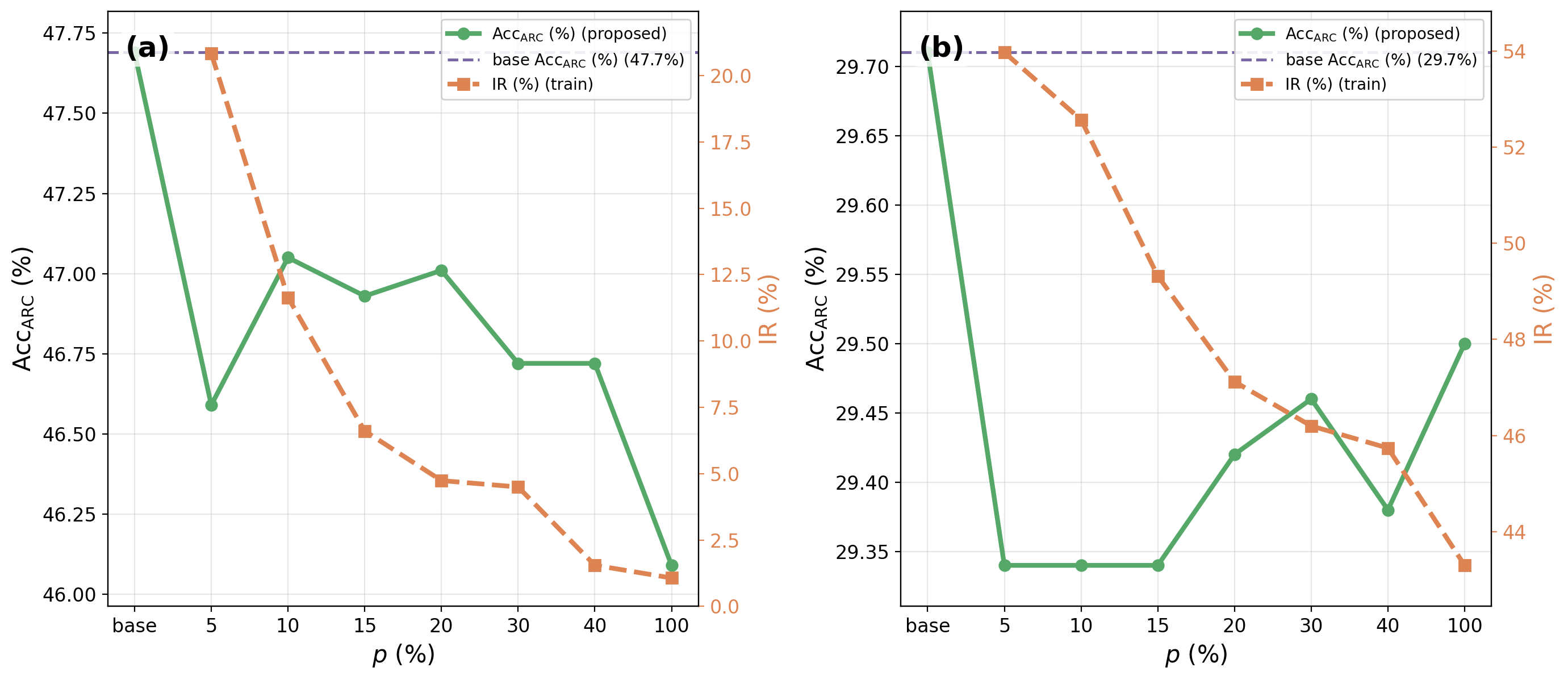} 
\caption{Comparison of $Acc_{ARC}$ and training $IR$ across varying values of $p$ for (a) \texttt{Qwen3.5-2B} and (b) \texttt{Gemma-4-E2B}.} 
\label{fig:capacity_vs_abliteration} 
\end{figure*}

\subsection{Downstream Utility and Task Capacity Preservation} 
\label{subsec:results_capacity} 
\noindent
This section investigates whether general model capacity is preserved following weight ablation.
Specifically, \texttt{ARC-Easy} \cite{pirtoaca2019answering} is employed to evaluate the reasoning and question-answering utility after ablation.
This benchmark assesses general question-answering and reasoning capabilities, making it well suited for compact architectures such as the 2B-parameter models studied in this paper.
As shown in Fig.~\ref{fig:capacity_vs_abliteration}, weight ablation induces negligible degradation in general question-answering performance. 
For \texttt{Qwen3.5-2B} in Fig.~\ref{fig:capacity_vs_abliteration}(a), \texttt{ARC-Easy} accuracy remains nearly invariant across all ablation levels, declining marginally from the unablated baseline of $47.7\%$ to $46.0\%$ even at $p=100\%$. 
Similarly, for \texttt{Gemma-4-E2B} in Fig.~\ref{fig:capacity_vs_abliteration}(b), \texttt{ARC-Easy} accuracy remains constant at $29.5\%$ across all thresholds up to $p=100\%$.
These findings confirm that the proposed method selectively removes self-identity representations without impairing core reasoning capabilities. 
Selecting an appropriate threshold $p$ thus enables the models to suppress self-identification while preserving general downstream utility.
The next experiment investigates if the factual knowledge is perserved after ablation.

\begin{figure*}
\centering 
\includegraphics[width=\textwidth]{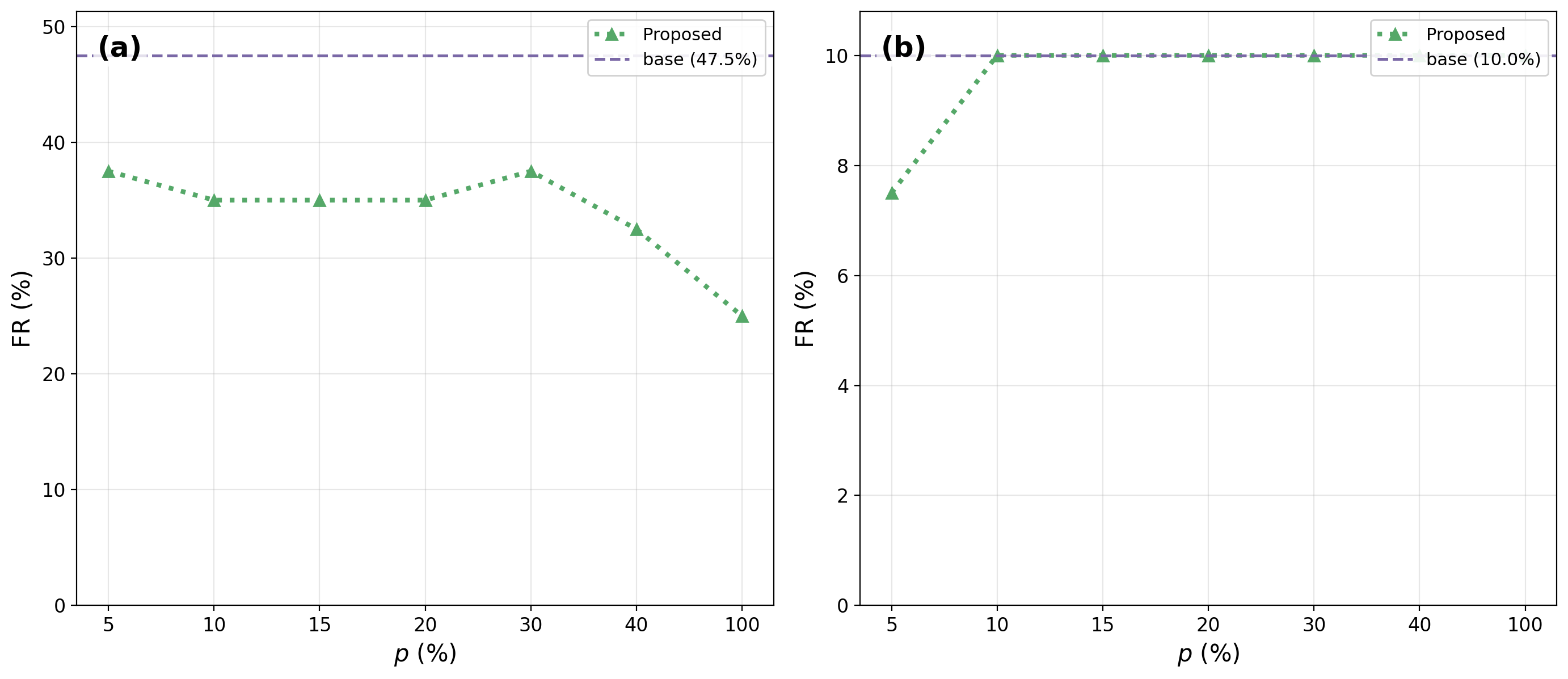} 
\caption{Effect of varying $p$ on $FR$ for (a) \texttt{Qwen3.5-2B} and (b) \texttt{Gemma-4-E2B}.} 
\label{fig:name_recall_self} 
\end{figure*}

\begin{figure*}
    \centering
    \includegraphics[width=\textwidth]{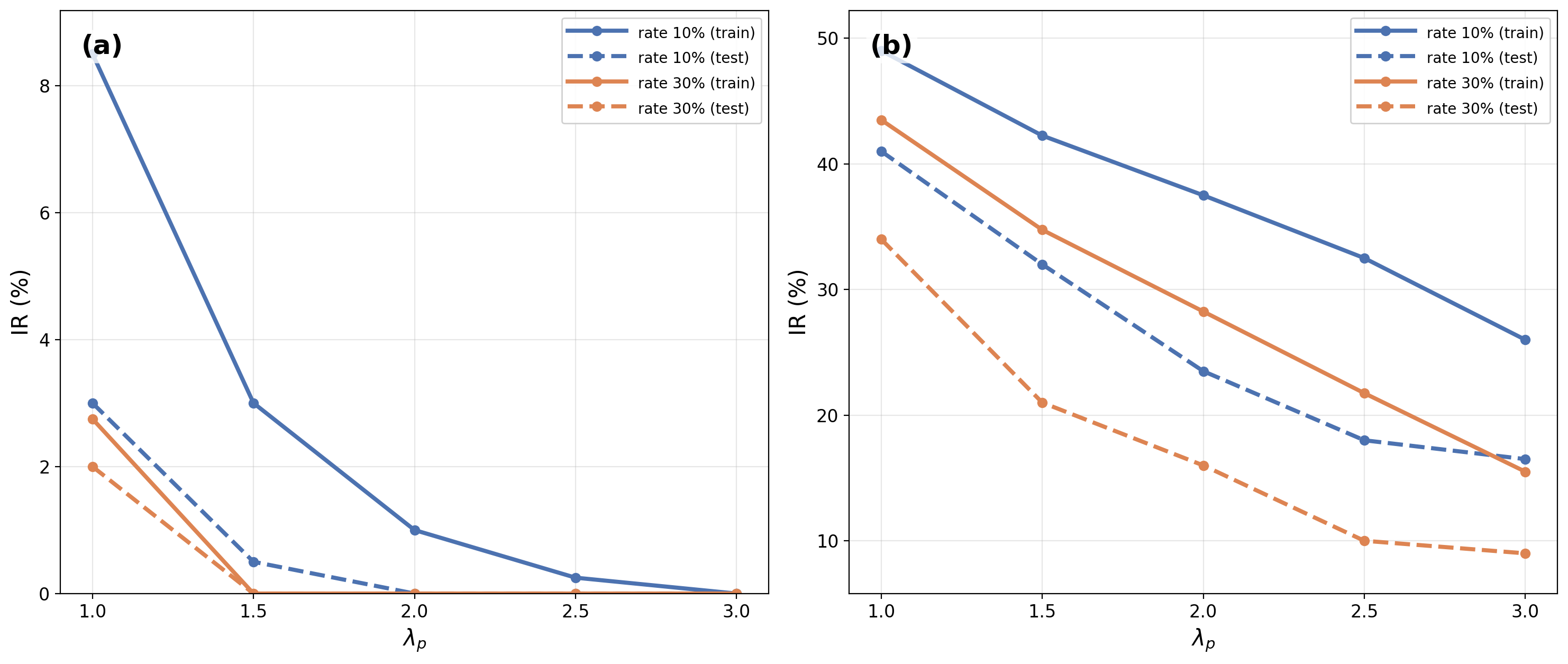}
    \caption{Effect of $\lambda_p$ on $IR$ for (a) \texttt{Qwen3.5-2B} and (b) \texttt{Gemma-4-E2B}.}
    \label{fig:fig7_rate_scale_force}
\end{figure*}

\subsection{Factual Knowledge Retrieval} 
\label{subsec:results_factual_recall} 
\noindent
A core requirement for effective machine unlearning is preserving general language modeling capabilities and objective factual knowledge. 
To verify that self-identity unlearning does not impair external factual retrieval, models are evaluated on objective queries referencing model names, such as ``Which open weight models are developed by Alibaba?''.
Fig.~\ref{fig:name_recall_self} illustrates $FR$ in factual contexts across varying values of $p$ relative to the base model. 
For \texttt{Qwen3.5-2B} in Fig.~\ref{fig:name_recall_self}(a), the base model exhibits a $47.5\%$ baseline name recall in factual responses. 
The \textit{Proposed} method gradually attenuates self-name occurrence to $25.0\%$ at $p=100\%$, indicating controlled suppression rather than catastrophic forgetting. 
For \texttt{Gemma-4-E2B} in Fig.~\ref{fig:name_recall_self}(b), baseline factual recall is lower at $10.0\%$. 
Under the proposed method, factual name recall remains stable at $10.0\%$ across all ablation thresholds $p \in [5\%, 100\%]$, aside from a minor fluctuation to $7.5\%$ at $p=5\%$. 
These results confirm that the proposed approach avoids collateral damage to objective factual retrieval involving target entity names at practical operating thresholds such as $p=30\%$.
The following experiment examines how the scaling factor $\lambda_p$ influences unlearning performance.

\subsection{Ablation Intensity} 
\noindent
Because the localized adapter weights encode self-identity representations, it is essential to examine how the ablation intensity $\lambda_p$---which scales the ablation magnitude applied to the top $p\%$ weight percentiles---affects unlearning performance.
Experiments are conducted to measure $IR$ across scaling factors $\lambda_p \in [1, 3]$ at percentile thresholds of $p=10\%$ and $p=30\%$.
Fig.~\ref{fig:fig7_rate_scale_force} presents the relationship between $\lambda_p$ and $IR$ for \texttt{Qwen3.5-2B} and \texttt{Gemma-4-E2B} across both training and testing splits. 
Overall, increasing $\lambda_p$ consistently reduces $IR$ across both architectures. 
However, \texttt{Gemma-4-E2B} exhibits substantially higher initial $IR$ values at lower scaling factors compared to \texttt{Qwen3.5-2B}, whereas \texttt{Qwen3.5-2B} drops rapidly to near-zero retention once $\lambda_p$ exceeds $1.5$. 
Across both subplots, the lower percentile setting ($10\%$) consistently maintains higher $IR$ than the higher setting ($30\%$), and training splits exhibit higher retention than testing splits across all evaluated scaling factors.
These results confirm that increasing $\lambda_p$ effectively amplifies the strength of weight ablation.

\subsection{Future Directions}
\noindent
Although these findings demonstrate the effectiveness of the proposed approach, several directions remain for future investigation. First, expanding empirical evaluations beyond compact architectures to larger language models is necessary to verify whether these trends hold at scale. A systematic analysis of key hyper-parameters, such as $\lambda_p$, $p$, and training epoch counts, will further clarify numerical stability, convergence behavior, and optimal operating regimes. Moreover, examining the functional properties of the identified weights through identity-swapping experiments will determine whether identical weight subsets govern both original and modified representations, formally testing the hypothesis that original identities are inherently more resistant to unlearning than newly acquired ones.
\\
\indent
Second, the scope of the methodology can be extended to more complex knowledge structures. The framework can be generalized from isolated entity names to higher-level factual knowledge and broader sensitive domains. Additionally, integrating the proposed method with prompt-based techniques offers a promising avenue for optimizing the retention-unlearning trade-off across diverse architectures, such as Gemma, and downstream applications. Together, these research directions will advance the understanding of representation dynamics and inform the design of more robust unlearning frameworks.

\section{Conclusion}
\label{sec:conclusion}
\noindent
This paper addresses the challenge of unintended model identity inheritance induced by training data sources, such as model distillation. 
It proposes a machine unlearning method that selectively ablates model self-identity without requiring extensive data filtering or exposing private client datasets.
By isolating identity-related weights within Low-Rank Adaptation (LoRA) adapters via targeted question-based fine-tuning and subsequently ablating high-magnitude adapter weights, the proposed method effectively removes self-identity representations.
Empirical evaluations on \texttt{Qwen3.5-2B} and \texttt{Gemma-4-E2B} demonstrate that the method reduces identity recall to near-zero levels for \texttt{Qwen3.5-2B} at $30\%$ weight ablation, substantially outperforming baseline strategies.
Furthermore, evaluations on the \texttt{ARC-Easy} benchmark and external factual recall tasks confirm that the proposed approach preserves core reasoning and general question-answering capabilities without degradation. 
Overall, this work provides a scalable, privacy-preserving framework for identity removal and parameter-efficient machine unlearning in open-weight large language models.

\bibliographystyle{IEEEtran}
\bibliography{id}
\end{document}